\documentclass{article}

\usepackage{iclr2027_conference,times}

\usepackage{amsmath}
\usepackage{amssymb}
\usepackage{algorithm}
\usepackage{algorithmic}
\usepackage{arydshln}
\usepackage{booktabs}
\usepackage{graphicx}
\usepackage{multirow}
\usepackage{subcaption}
\usepackage[table]{xcolor}
\usepackage{url}
\usepackage{hyperref}

\definecolor{deltaGreen}{HTML}{16856F}

\newcommand{\resizetable}[2][1.0]{%
    \resizebox{#1\linewidth}{!}{#2}%
}
\newcommand{\deltatext}[1]{%
    {\footnotesize\color{deltaGreen}\itshape #1\/}%
}
\newcommand{\runinheading}[1]{%
    \subsubsubsection{\textbf{#1}}%
}

\iclrfinalcopy

\title{ADOPD: Reference-Privileged On-Policy Distillation for MLLM-Based Industrial \\ Anomaly Detection}

\author{%
    Jingtai He, Shiyuan Meng, Wenchao Meng\thanks{corresponding author}, Qinmin Yang\\
    Zhejiang University\\
    \texttt{\{heatai,wmengzju\}@zju.edu.cn}%
}

\begin{document}

\maketitle
\fancyhead{}

\begin{abstract}
Industrial anomaly detection (IAD) requires identifying fine-grained deviations from normal visual patterns. Multimodal large language models (MLLMs) can improve recognition accuracy by comparing query images with references at inference time, but these benefits rely on additional retrieval and processing. We investigate whether the benefits of reference comparison can instead be internalized in the model parameters. Access to references during training allows a reference-aware teacher to supervise a query-only student. However, the teacher may favor plausible responses based on query cues or language priors rather than valid visual information. We propose ADOPD, a reference-privileged on-policy distillation framework. The teacher evaluates student-generated rollouts under matched and mismatched references. The matched-reference teacher-to-student log-ratio defines the token-level learning direction, specifying what the student should learn. The likelihood gap between the two reference views estimates reference-specific support and calibrates the sequence-level weight. ADOPD achieves 77.31\% average accuracy on the MMAD benchmark under zero-shot inference, improving the Qwen3-VL-4B backbone by 6.14 points and outperforming its one-shot setting by 2.64 points. Experiments show that ADOPD learns a fine-grained anomaly inspection strategy from reference comparison. The project will be available at \url{https://github.com/withTai/ADOPD}.

\end{abstract}


\section{Introduction}
\label{sec:introduction}

Industrial anomaly detection (IAD) is a fundamental task in automated manufacturing and quality control, aiming to identify defects or irregularities that deviate from normal patterns in complex visual data~\citep{visa,anomalyclip,moead}. In real-world applications, industrial inspection increasingly requires open-set recognition and explainable reasoning about defect type, location, and appearance. Multimodal large language models (MLLMs) offer a promising interface for these requirements because they combine visual perception with flexible language reasoning~\citep{anomalygpt,anomalyr1}. However, their open-domain representations struggle to capture fine-grained local deviations and product-specific anomaly patterns.

\begin{figure}[t]
    \centering
    \includegraphics[width=0.9\columnwidth]{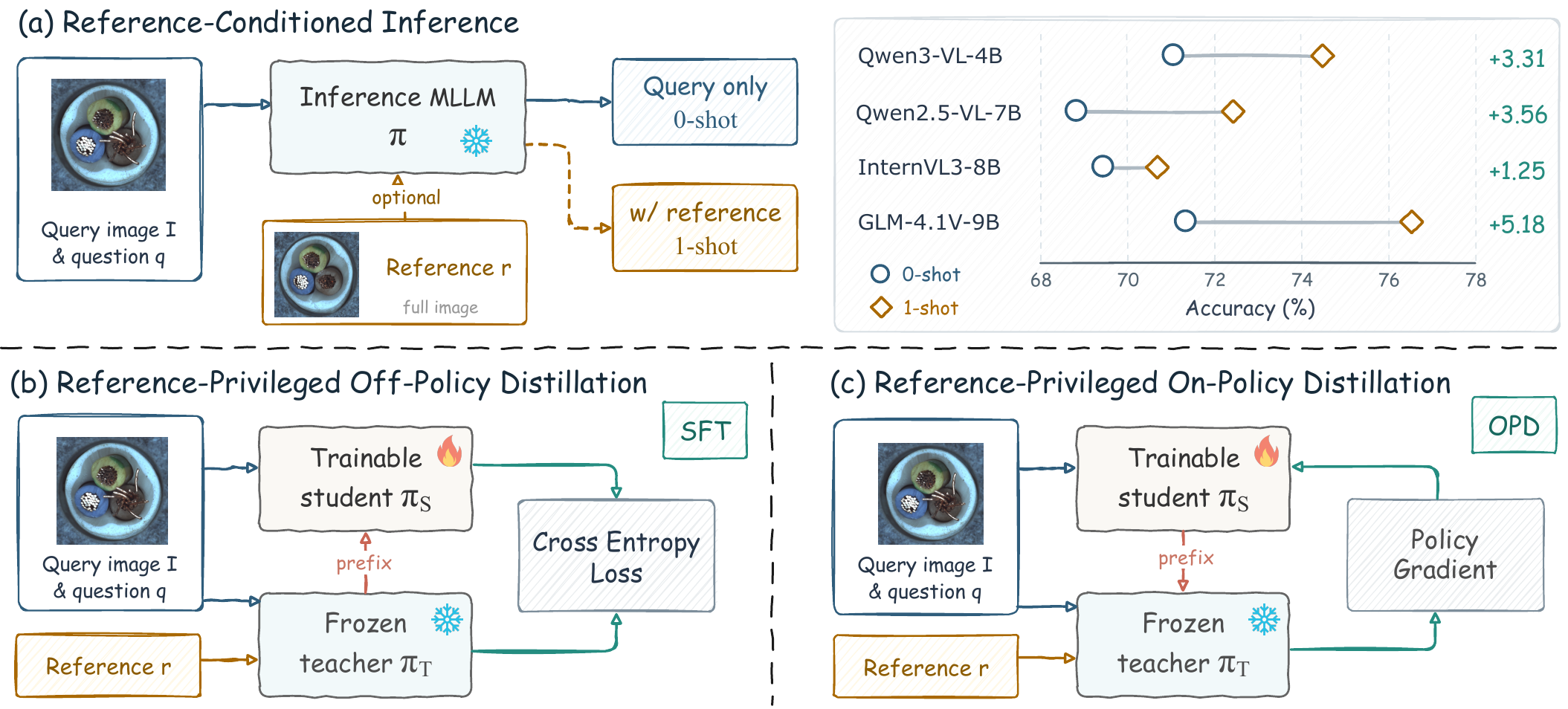}
    \caption{\textbf{Reference-conditioned inference to reference-privileged training.} (a) A normal reference improves all four tested MLLM backbones. (b) Reference-privileged off-policy distillation trains the student by minimizing the cross-entropy loss on fixed teacher-generated targets. (c) Reference-privileged on-policy distillation evaluates student-generated prefixes with the reference-aware teacher and updates the student through a policy-gradient objective.}
    \label{fig:teaser}
\end{figure}

Reference comparison supplies the missing product-specific visual context. Similar to human inspectors comparing products against standard templates, reference-conditioned MLLMs can use regional examples to identify fine-grained deviations that are ambiguous from the query alone~\citep{patchcore,mmad}. As Figure~\ref{fig:teaser}(a) shows, one-shot reference-conditioned inference improves over zero-shot inference, indicating that reference comparison provides useful evidence for MLLM-based IAD. Nevertheless, explicit reference-conditioned inference requires suitable images to be retrieved and processed for every query, introducing additional computational overhead. We therefore ask: \textbf{\textit{Can the benefits of reference comparison be internalized into model parameters during training to improve MLLM performance in IAD?}}

Training-time reference access naturally fits the privileged-information setting \citep{privileged}. A teacher can observe both the query and reference evidence, whereas the student observes only the query, thereby creating reference-privileged supervision signals. As Figure~\ref{fig:teaser}(b) shows, existing methods~\citep{anomalygpt,adcopilot} use supervised fine-tuning (SFT) to distill reference-conditioned teacher behavior into a query-only student through long chains of thought. However, SFT is an off-policy distillation method that trains the student on fixed teacher-generated targets rather than on states induced by the student itself. This off-policy approach can cause exposure bias, where the student fails to generalize from its own generation trajectories at inference.

On-policy distillation (OPD) addresses this mismatch by evaluating teacher feedback on student-generated prefixes~\citep{thinkingmachine,revisiting}. Figure~\ref{fig:teaser}(c) shows that the student first samples a rollout and the reference-aware teacher then evaluates the student-generated prefixes, inducing a policy-gradient update from the teacher-student log-ratio.
However, OPD alone does not establish that the teacher's support is grounded in the intended comparison. Even under a valid reference, the teacher may favor a plausible response because of query-image cues or language priors that are insensitive to the query-reference relation. The remaining challenge is therefore to determine how strongly reference-aware feedback on each student rollout should be trusted.

We propose ADOPD, a reference-privileged on-policy distillation framework that separates \textit{what to learn} from \textit{how strongly to learn it}. The student first samples rollouts from the query-only prompt, so every prefix is induced by the current student. A frozen teacher then replays the same rollout tokens under two asymmetric reference views: a matched regional reference that preserves the valid query-reference relation, and a mismatched regional reference that deliberately breaks it. The matched-reference teacher-to-student log-probability ratio defines the token-level OPD direction, specifying what the student should learn. The matched-mismatched teacher likelihood gap yields a group-relative sequence weight, specifying how strongly the rollout should contribute. ADOPD uses the reference comparison to teach the model a fine-grained anomaly inspection strategy.

We conduct extensive experiments to validate the effectiveness of ADOPD. With only 6K constructed paired training records, ADOPD achieves 77.31\% average accuracy on the MMAD~\citep{mmad} benchmark under zero-shot inference. This improves the Qwen3-VL-4B backbone by 6.14 points and exceeds the reference-conditioned one-shot counterpart by 2.64 points. It also outperforms the SFT counterpart by 1.61 points without additional long reasoning chains. These results indicate that ADOPD learns anomaly inspection behavior shaped by reference comparison, improving MLLM performance in IAD.

Our contributions are summarized as follows:
\begin{itemize}
    \item We propose ADOPD, a reference-privileged on-policy distillation framework that internalizes the benefits of inference-time reference comparison into model parameters.
    \item We introduce an asymmetric matched/mismatched reference intervention: the matched view defines the token-level OPD direction, while the mismatched view control estimates sequence-level update strength.
    \item Experiments show that ADOPD learns anomaly inspection behavior shaped by reference comparison, improving MLLM performance in IAD.
\end{itemize}


\section{Preliminaries: On-Policy Distillation}
\label{sec:preliminary}

Knowledge distillation transfers the behavior of a strong teacher $\pi_T$ to a student $\pi_S$ by aligning their output distributions. On-policy distillation (OPD) evaluates the teacher on prefixes generated by the student~\citep{thinkingmachine,revisiting}. Given the prompt $x$, OPD minimizes the reverse KL divergence between the student and the teacher on trajectories sampled from the student:
\begin{equation}
    \mathcal{J}_{\mathrm{OPD}}
    =\mathbb{E}_{x \sim \mathcal{D}}\big[ D_{\mathrm{KL}} \big(\pi_S (\cdot \mid x) \parallel \pi_T(\cdot \mid x)\big) \big].
    \label{eq:opd_local_reverse_kl}
\end{equation}
Differentiating this optimization objective with respect to the student parameters $\theta$ gives\footnote{Detailed derivations are in Appendix~\ref{app:opd_derivation}.}
\begin{equation}
    \nabla_{\theta} \mathcal{J}_{\mathrm{OPD}} =
    \mathbb{E}_{x \sim \mathcal{D}, y \sim \pi_S(\cdot \mid x)}
    \left[ \log \frac{\pi_{S}(y \mid x)}{\pi_T(y \mid x)}
    \nabla_{\theta} \log \pi_{S}(y \mid x)
    \right].
    \label{eq:opd_local_reverse_kl_gradient}
\end{equation}
Practical sampled-token OPD adopts a lower-variance local approximation by treating each student-generated prefix as stop-gradient states. Thus, evaluating the remaining term at a student-sampled token $y_t$ provides an unbiased Monte Carlo estimate of the fixed-state local gradient. As in RL, the negative of its reverse-KL coefficient defines a dense token-level advantage~\citep{sdpo},
\begin{equation}
    A_t
    =\operatorname{sg}\!\left[
        \log \pi_T(y_t \mid x, y_{<t}) - \log \pi_S(y_t\mid x, y_{<t})
    \right],
    \label{eq:opd_token_advantage}
\end{equation}
where $\operatorname{sg}$ stops gradients through the advantage. This advantage is positive when the teacher assigns more probability to the sampled token than the student, and negative when the student overweights that token. The corresponding stop-gradient training surrogate is
\begin{equation}
    \widehat{\mathcal{L}}_{\mathrm{OPD}}
    =-\frac{1}{|y|}\sum_{t=1}^{|y|}
    A_t\log\pi_S(y_t\mid x,y_{<t}).
    \label{eq:pg_style_opd}
\end{equation}
Practical sampled-token OPD treats student-induced prefixes as stop-gradient training states. Its estimator is unbiased for the conditional reverse-KL gradient at a fixed prefix, but does not backpropagate through how earlier tokens alter future visited states.


\section{Method}
\label{sec:method}

\begin{figure*}[t]
    \centering
    \includegraphics[width=\textwidth]{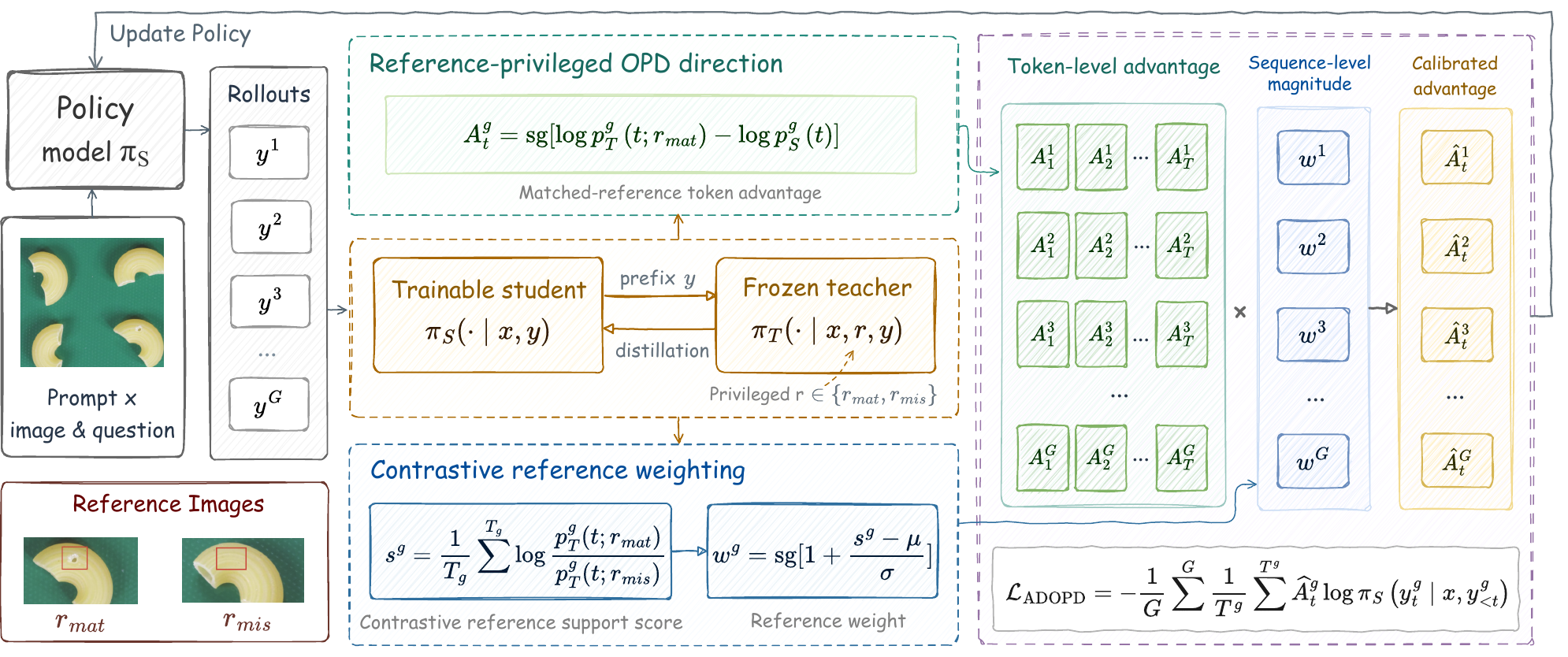}
    \caption{\textbf{Overview of ADOPD.} For each prompt $x$, the student $\pi_S$ samples $G$ rollouts. A frozen teacher $\pi_T$ evaluates the same tokens with matched and mismatched references. The matched-reference teacher--student log-probability ratio gives the token-level advantage $A_t^g$. The teacher likelihood ratio between the two reference views gives the sequence-level magnitude $w^g$ to form $\widehat{A}_t^g$.}
    \label{fig:method_pipeline}
\end{figure*}

ADOPD internalizes the benefits of inference-time reference comparison by separating \emph{what to learn} from \emph{how strongly to learn it}. As shown in Figure~\ref{fig:method_pipeline}, we design reference-privileged on-policy distillation and contrastive reference weighting mechanisms to achieve token-level direction and sequence-level magnitude, respectively. During this stage, only the student policy is optimized.

\subsection{Reference-Privileged On-Policy Distillation}
\label{sec:problem_setting}

Let $x$ contain the inspection question and the original query image. The trainable policy $\pi_S$ samples $G$ sibling rollouts:
\begin{equation}
    y^{g} \sim \pi_S(\cdot\mid x),
    \qquad g=1,\ldots,G.
    \label{eq:student_rollouts}
\end{equation}
Here, $y_{<t}^g$ denotes the prefix of rollout $g$ at token $t$. Because the rollout is sampled by the current student, each prefix is an on-policy training state. At generation step $t$, both the student and teacher are evaluated along the same student-generated prefix. The two next-token distributions are
\begin{align}
    p_S^g(t) &:= \pi_S(y_t^g\mid x,y_{<t}^g),\\
    p_T^g(t;r) &:= \pi_T(y_t^g\mid x,r,y_{<t}^g).
\end{align}
For each rollout, the frozen teacher $\pi_T$ receives either a matched reference $r_{\mathrm{mat}}$ or a mismatched reference $r_{\mathrm{mis}}$. The matched reference is semantically consistent with the query anomaly state and is the only candidate supervision source. The mismatched reference contains regional evidence from the opposite anomaly state and deliberately breaks the valid query--reference relation. Thus, \emph{matched} and \emph{mismatched} describe semantic consistency with the query rather than fixed normal and defective labels; the two views are asymmetric and cannot exchange roles.

The matched-reference teacher defines what the student should learn. Holding the sampled token and student-generated prefix fixed, the matched-reference view instantiates the OPD token advantage with privileged conditioning:
\begin{equation}
    A_t^g=\operatorname{sg}\!\left[
        \log p_T^g(t;r_{\mathrm{mat}})-\log p_S^g(t)
    \right].
    \label{eq:matched_reference_advantage}
\end{equation}
The matched-reference token-level advantage $A_t^g$ determines the distillation direction. It reinforces a token that the matched-reference teacher prefers more than the student does, and suppresses tokens that the student overweights. It provides a candidate distillation direction but does not by itself determine whether the teacher's support is genuinely grounded in the comparison.

\subsection{Contrastive Reference Weighting}
\label{sec:reference_weighting}

Matched-reference confidence alone might be ambiguous because the teacher can rely on the query image or its language prior without using the intended reference comparison. ADOPD tests the reference specificity of each rollout by replaying the same trajectory under the mismatched-reference control. The teacher log-probability gap between the matched and mismatched reference views defines the contrastive reference support score:
\begin{equation}
    \Delta_t^g =
        \log p_T^g(t;r_{\mathrm{mat}})
        -\log p_T^g(t;r_{\mathrm{mis}}),
    \qquad
    s^g=\frac{1}{T_g}\sum_{t=1}^{T_g}\Delta_t^g,
    \label{eq:reference_support_score}
\end{equation}
where $T_g$ is the length of rollout $g$. Mathematically, $s^{g}$ is a reference-view likelihood asymmetry evaluated on a single student trajectory; semantically, $s^{g}$ indicates whether that trajectory is preferentially supported when the valid reference relation is preserved. We aggregate over the sequence because single-token gaps are dominated by syntax and generic reasoning, whereas a diagnosis spans many tokens and one sequence-level magnitude keeps the resulting update coherent. As no answer is involved, $s^{g}$ is a verifier-free proxy for how strongly matched-view supervision should be updated.

Raw support scores vary across queries and product categories. Contrastive reference weighting therefore calibrates $s^{g}$ against the mean $\mu$ and standard deviation $\sigma$ of $\left\{s^g\right\}_{g=1}^{G}$ for the same prompt, producing the group-relative reference weight:
\begin{equation}
    w^{g} = \operatorname{sg}\!\left[
        1 + \operatorname{clip}\!\left(
            \frac{s^{g}-\mu}{\sigma},
            -\epsilon_w,\epsilon_w
        \right)
    \right],
    \label{eq:rollout_weight}
\end{equation}
where $\epsilon_w$ determines the symmetric clipping bounds. The unit offset retains standard OPD strength as the neutral scale, while clipping prevents a few trajectories from dominating the update. Since the weight $w^{g}$ is detached and non-negative, the mismatched view control rescales the learning magnitude but never reverses the matched-reference OPD direction.

\begin{algorithm*}[!t]
\caption{ADOPD}
\label{alg:adopd}
\begin{algorithmic}[1]
\setlength{\itemsep}{2pt}
\REQUIRE Training triples $(x,r_{\mathrm{mat}},r_{\mathrm{mis}})$, student $\pi_S$, teacher $\pi_T$, group size $G$, clip bounds $\epsilon_w$
\FOR{each training iteration}
    \FOR{each prompt $x$ with privileged information $r_{\mathrm{mat}}$ and $r_{\mathrm{mis}}$}
        \STATE Sample $\{y^1, y^2,\dots, y^G\}\sim\pi_S(\cdot\mid x)$ \COMMENT{student-induced states}
        \FOR{$g=1,\ldots,G$}
            \FOR{$t=1,\ldots,T_g$}
                \vspace{0.25em}
                \STATE $A_t^g=\operatorname{sg}\!\left[\log p_T^g(t;r_{\mathrm{mat}})-\log p_S^g(t)\right]$ \COMMENT{token-level OPD direction}
                \vspace{0.35em}
                \STATE $\Delta_t^g =\log p_T^g(t;r_{\mathrm{mat}})-\log p_T^g(t;r_{\mathrm{mis}})$
                \vspace{0.1em}
            \ENDFOR
            \STATE $s^g\leftarrow\frac{1}{T_g}\sum_{t=1}^{T_g}\Delta_t^g$ \COMMENT{contrastive reference support score}
        \ENDFOR
        \STATE $\mu,\sigma\leftarrow\operatorname{mean/std}(\left\{s^g\right\}_{g=1}^{G})$
        \FOR{$g=1,\ldots,G$}
            \STATE $w^g\leftarrow\operatorname{sg}\!\left[1+\operatorname{clip}\!\left((s^g-\mu)/\sigma,-\epsilon_w,\epsilon_w\right)\right]$ \COMMENT{group-relative reference weight}
            \STATE $\widehat{A}_t^g\leftarrow w^g A_t^g$ for all $t$ in $y^g$ \COMMENT{support-calibrated token advantage}
        \ENDFOR
    \ENDFOR
    \STATE Update $\pi_S$ with $\mathcal{L}_{\mathrm{ADOPD}}=-\frac{1}{G}\sum_{g=1}^{G}\frac{1}{T_g}\sum_{t=1}^{T_g}\widehat{A}_t^g\log\pi_S(y_t^g\mid x,y_{<t}^g)$
\ENDFOR
\end{algorithmic}
\end{algorithm*}

\subsection{Calibrated Optimization Objective}
\label{sec:training_inference}
Broadcasting the sequence-level weight $w^{g}$ to every token in rollout $g$ yields the support-calibrated token advantage:
\begin{equation}
    \widehat{A}_t^{g}=w^{g}A_t^{g}.
    \label{eq:calibration}
\end{equation}

$\widehat{A}_t^g$ combines the matched-reference token-level direction with the contrastive reference sequence-level magnitude, providing a calibrated advantage for policy-gradient updates. The resulting ADOPD objective for the student policy is
\begin{equation}
    \mathcal{L}_{\mathrm{ADOPD}}
    =-\frac{1}{G}\sum_{g=1}^{G}\frac{1}{T_g}
    \sum_{t=1}^{T_g}\widehat{A}_t^{g}
    \log\pi_S(y_t^{g}\mid x,y_{<t}^{g}).
    \label{eq:adopd_objective}
\end{equation}


\section{Experiments}
\label{sec:experiments}

\subsection{Experiment Setup}
\label{sec:experiment_setup}

\runinheading{Evaluation Protocol.}
We evaluate ADOPD on the comprehensive industrial anomaly detection benchmark MMAD~\citep{mmad}, which aggregates MVTec-AD~\citep{mvtec}, VisA~\citep{visa}, MVTec-LOCO~\citep{MVTECLOCO}, and GoodsAD~\citep{goodsad}. It contains seven subtasks: anomaly discrimination; defect classification, localization, description, and analysis; and object classification and analysis. We report accuracy for each subtask and the arithmetic mean across subtasks. Unless stated otherwise, zero-shot evaluation supplies no reference image at inference, whereas one-shot evaluation supplies one normal reference image.

\runinheading{Baselines.}
We compare ADOPD with open-source MLLMs from two groups: open-source models and IAD-specific models. The open-source group includes LLaVA-1.5 and LLaVA-NeXT~\citep{llava,llavanext}, Qwen2.5-VL and Qwen3-VL~\citep{qwen2.5vl,qwen3vl}, InternVL3~\citep{internvl3}, and GLM-4.1V~\citep{glm}. The domain-specific group includes AnomalyR1~\citep{anomalyr1}, IAD-R1~\citep{iadr1}, and Reason-IAD~\citep{reasoniad}.

\runinheading{Training Dataset.}
We construct 6,000 training records from Real-IAD~\citep{realiad}, covering 30 product categories and five camera views. Note that there is no overlap between the training set and the evaluation benchmarks. Each defective image is paired with a normal image from the same product, instance, and camera. Appendix~\ref{app:realiad_paired_dataset} details the dataset construction. The student receives only the original image and inspection question. During training, the frozen teacher additionally receives matched and mismatched regional crops derived from the image pair and defect annotation. These regional annotations and reference images are privileged training information and are unavailable during evaluation.

\runinheading{Implementation Details.}
Our default student and teacher are Qwen3-VL-4B and Qwen3-VL-32B, respectively. We implement ADOPD in VeRL~\citep{hybridflow} and train for 60 optimization steps on a node equipped with NVIDIA RTX PRO 6000 GPUs. The training batch size is 64, the mini-batch size is 16, the number of rollouts per prompt is $G=8$, and the learning rate is $1\times10^{-6}$. We use VeRL's sampled-token reverse-KL estimator for the matched-reference token advantage and clip the group-relative reference weight to $\epsilon_w=1$.

\begin{table}[t]
\caption{\textbf{Comparison of MLLMs on the MMAD benchmark.} All methods are evaluated in the zero-shot setting. Best and second-best results are shown in bold and underlined, respectively.}
\label{table:0shot}
\centering
{
    \renewcommand{\arraystretch}{1.07}
    \resizetable{%
    \begin{tabular}{lc*{8}{c}}
        \toprule
        \multirow{2}{*}{\textbf{Model}}
        & \multirow{2}{*}{\textbf{Scale}}
        & \multicolumn{1}{c}{\textbf{Anomaly}}
        & \multicolumn{4}{c}{\textbf{Defect}}
        & \multicolumn{2}{c}{\textbf{Object}}
        & \multirow{2}{*}{\textbf{Average}} \\
        \cmidrule(lr){3-3}\cmidrule(lr){4-7}\cmidrule(lr){8-9}
        & & \textbf{Discrimination} & \textbf{Classification} & \textbf{Localization}
        & \textbf{Description} & \textbf{Analysis} & \textbf{Classification} & \textbf{Analysis} & \\
        \midrule
        \multicolumn{10}{l}{\textit{Open-Source MLLMs}} \\
        \addlinespace[2pt]
        Qwen3-VL     & 2B  & 49.80 & 48.77 & 52.33          & 70.00          & 80.04          & 92.58          & 81.30          & 67.83 \\
        Qwen2.5-VL   & 3B  & 60.63 & 47.03 & 54.44          & 63.51          & 79.25          & 87.48          & 82.02          & 67.77 \\
        Qwen3-VL     & 4B  & 65.87 & 53.90 & 58.58          & 66.40 & 77.93          & 92.13 & 83.37 & 71.17 \\
        Qwen2.5-VL   & 7B  & 60.16 & 49.94 & 57.43          & 60.37 & 75.51          & 93.43 & \underline{84.90} & 68.82 \\
        LLaVA-1.5    & 7B  & 54.01 & 38.22 & 40.59          & 50.43 & 68.74          & 70.74 & 70.58 & 56.19 \\
        InternVL3    & 8B  & 60.88 & 50.57 & 59.99          & 66.33 & 76.90          & 86.67 & 84.80 & 69.45 \\
        Qwen3-VL     & 8B  & 67.20 & \underline{58.15} & 58.74 & 66.88 & 77.40 & 92.70 & 83.74 & 72.12 \\
        LLaVA-NeXT   & 8B  & 57.91 & 43.15 & 49.21          & 61.56 & 75.60          & 71.56 & 72.33 & 61.62 \\
        GLM-4.1V     & 9B  & 61.55 & 52.90 & 59.21          & 65.85 & 81.63 & \underline{93.81} & 84.65 & 71.37 \\
        \midrule
        \multicolumn{10}{l}{\textit{Domain-Specific MLLMs}} \\
        \addlinespace[2pt]
        AnomalyR1    & 3B & 75.34 & 45.49 & \underline{61.15} & 68.96 & \underline{83.88} & 90.30 & \textbf{84.99} & 72.87 \\
        IAD-R1       & 7B & \underline{75.72} & 54.53 & 60.72 & 67.43 & 77.65 & 93.17 & 84.02 & 73.32 \\
        Reason-IAD   & 8B & 67.73 & \textbf{73.12} & 59.74 & \textbf{74.69} & 80.65 & \textbf{96.78} & 84.24 & \underline{76.71} \\
        \addlinespace[2pt]
        \rowcolor{gray!8}
        ADOPD & 4B & \textbf{88.44} & 57.03 & \textbf{63.64} & \underline{70.78} & \textbf{84.16} & 92.77 & 84.32 & \textbf{77.31} \\
        \bottomrule
    \end{tabular}
    }
}
\end{table}

\subsection{Main Results}

\runinheading{Industrial Anomaly Detection Accuracy.}
Table~\ref{table:0shot} reports the main zero-shot comparison. ADOPD obtains the highest mean accuracy among the evaluated open-source models, reaching 77.31\%. ADOPD improves the average accuracy of its Qwen3-VL-4B backbone by 6.14 percentage points without accessing reference images at inference. It is also higher than the strongest listed IAD-specific model, Reason-IAD, by 0.60 points while using a 4B rather than an 8B backbone. The per-subtask scores identify where the gain occurs. Relative to the best competing score in each subtask, ADOPD improves anomaly discrimination by 12.72 points, defect localization by 2.49 points, and defect analysis by 0.28 points. These tasks require determining whether and where an observation differs from normality. In contrast, ADOPD underperforms the best competitor on defect classification and description, object classification and analysis. The improvement is concentrated in comparison-sensitive anomaly recognition rather than distributed across all semantic reasoning tasks. The results support the hypothesis that ADOPD utilizes the reference comparison to teach the model a fine-grained anomaly inspection strategy.

\begin{figure}[t]
    \centering
    \begin{subfigure}[t]{0.47\linewidth}
        \centering
        \includegraphics[width=\linewidth]{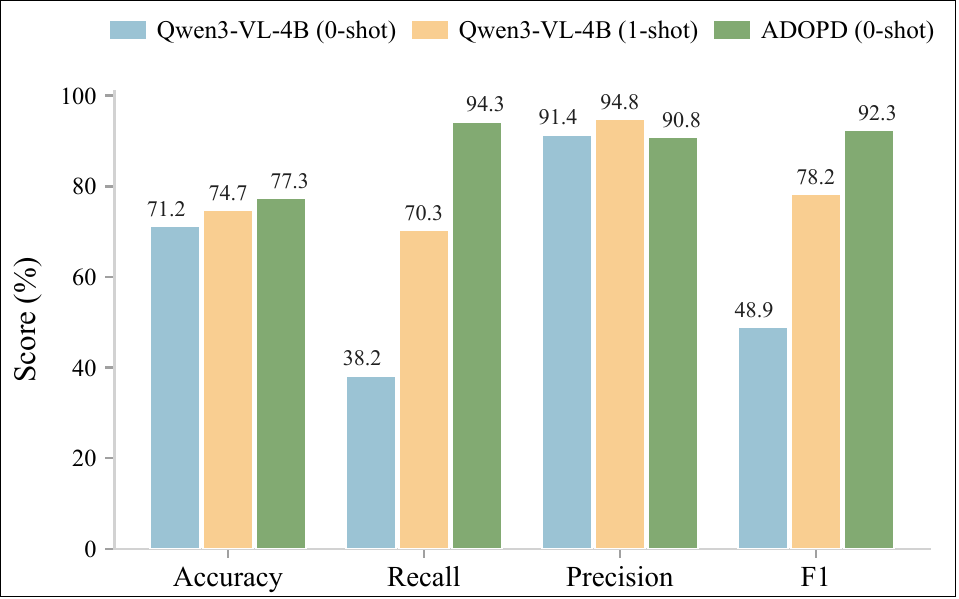}
        \caption{Aggregate MMAD metrics.}
        \label{fig:experiment_1}
    \end{subfigure}
    \hfill
    \begin{subfigure}[t]{0.47\linewidth}
        \centering
        \includegraphics[width=\linewidth]{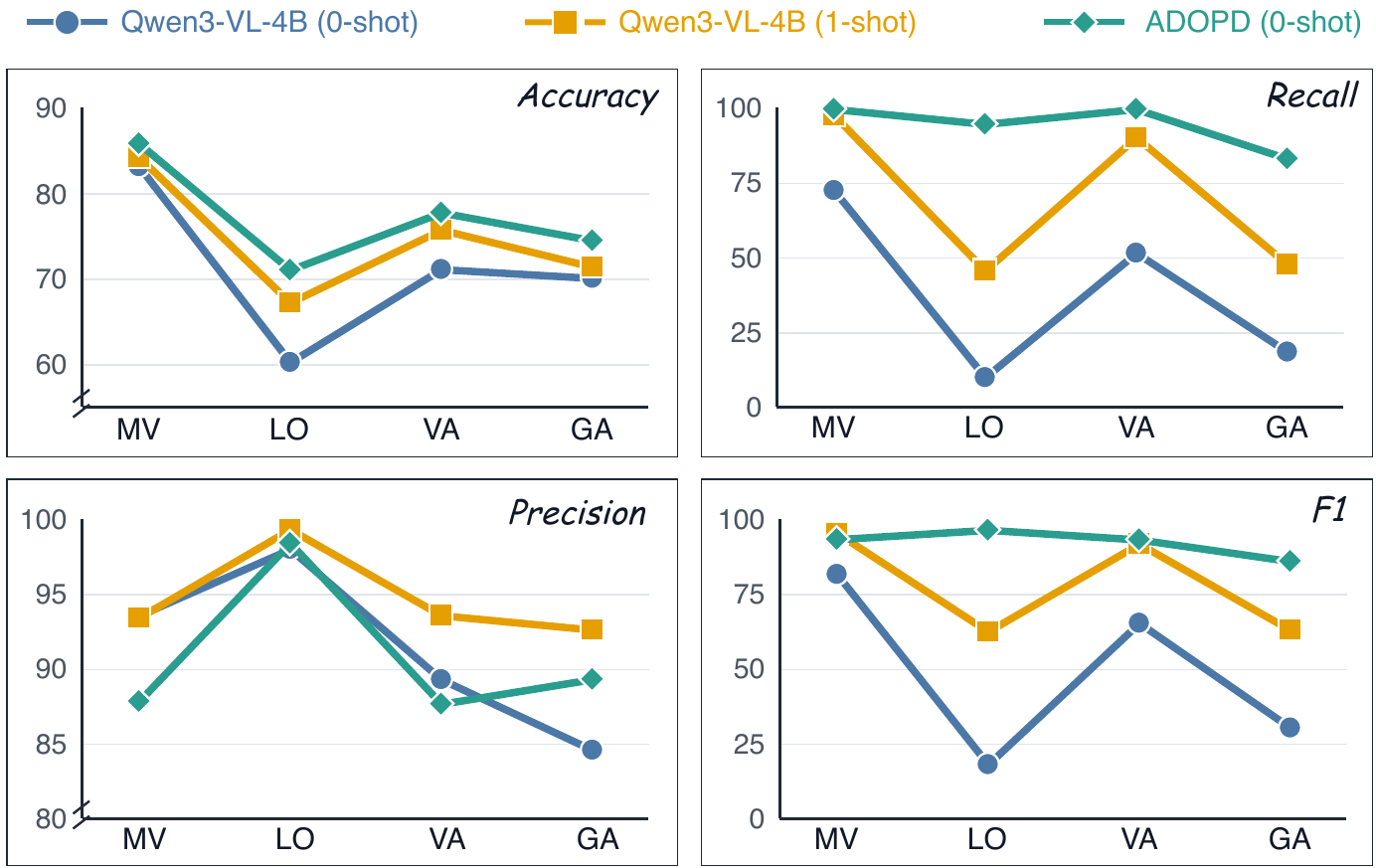}
        \caption{Dataset-wise MMAD metrics.}
        \label{fig:experiment_2}
    \end{subfigure}
    \caption{\textbf{MMAD results for ADOPD and Qwen3-VL-4B baselines.} (a) Aggregate accuracy, recall, precision, and F1 under zero-shot and one-shot inference. (b) Dataset-wise accuracy, recall, precision, and F1 under zero-shot and one-shot inference.}
    \label{fig:main_results}
\end{figure}

\runinheading{Retaining the Benefits of Reference Comparison.}
Figure~\ref{fig:main_results}(\subref{fig:experiment_1}) confirms that explicit reference-conditioned inference helps the backbone: moving Qwen3-VL-4B from zero-shot to one-shot improves all reported metrics. In IAD, recall measures how many defective samples are detected and therefore reflects missed defects; precision measures how many predicted anomalies are genuine and therefore reflects false alarms. Despite receiving no reference at inference, ADOPD reaches 77.31\% accuracy, 94.27\% recall, and 92.32\% F1. Compared with the one-shot backbone, this corresponds to gains of 2.64, 23.95, and 14.08 points, respectively, while reducing precision by 3.93 points. As shown in Figure~\ref{fig:main_results}(\subref{fig:experiment_2}), ADOPD improves accuracy over the zero-shot backbone and is higher than the one-shot backbone on every dataset. Recall increases consistently, whereas precision varies more across datasets, which explains why F1 shows the largest aggregate gain. Together, these results show that ADOPD achieves higher defect sensitivity and fewer missed anomalies than both the zero-shot and one-shot baselines, retaining the benefits of reference comparison at inference.

\subsection{Ablation Studies}

\runinheading{Training Paradigm.}
We isolate the effect of the training objective by replacing ADOPD with two paradigms under the same data and model configuration. One paradigm is off-policy SFT, which trains the query-only student using cross-entropy on fixed responses generated by the reference-privileged teacher. The other paradigm is reference-privileged OPD, which evaluates the teacher on student-generated prefixes but applies no contrastive reference weighting. Table~\ref{tab:paradigm} shows that SFT and reference-privileged OPD reach 75.70\% and 74.95\% average accuracy, improving the baseline by 4.53 and 3.78 points, respectively. However, SFT's advantage over uncalibrated OPD indicates that on-policy sampling alone does not guarantee stronger transfer. Calibrated ADOPD achieves the best average accuracy in this ablation, showing that contrastive reference weighting is important for retaining the benefits of reference comparison.

\begin{table}[t]
    \caption{\textbf{Effect of training paradigms.} We compare SFT, Reference-Privileged OPD, and calibrated ADOPD on the MMAD subsets. Results are reported as accuracy (\%) in the zero-shot setting.}
    \label{tab:paradigm}
    \centering
    \renewcommand{\arraystretch}{1.15}
    \resizetable[0.89]{
        \begin{tabular}{lccccc}
            \toprule
            Training Paradigm & MVTec-AD & MVTec-LOCO & VisA & GoodsAD & Average \\
            \midrule
            Base Model      & 83.17    & 60.27   & 71.13   & 70.11     & 71.17  \\
            SFT (off-policy distillation)      & 85.98    & 67.89   & 76.55   & 72.38     & 75.70  \\
            Reference-Privileged OPD    & \textbf{86.15}   & 66.08    & 75.72   & 71.86    & 74.95 \\
            ADOPD (w/ calibration) & 85.86 & \textbf{71.07} & \textbf{77.78} & \textbf{74.52} & \textbf{77.31} \\
            \bottomrule
        \end{tabular}
    }
\end{table}

\runinheading{Reference Granularity.}
Table~\ref{tab:construction} studies the effect of the granularity of privileged references. The full-image configuration reaches 74.67\% average accuracy, whereas all configurations using regional crops are above 75\%, showing a clear performance advantage. The granularity of privileged references determines whether the teacher can provide supervision that is specific to a local anomaly rather than to the whole product. These results indicate that the benefit is not merely from supplying an additional reference image: constraining privileged supervision to the anomaly-relevant region produces a more effective signal for fine-grained visual comparison.

\begin{table}[t]
    \caption{\textbf{Effect of reference construction.} We compare different reference granularities and contrastive setups on the MMAD subsets. Results are reported as accuracy (\%) in the zero-shot setting.}
    \label{tab:construction}
    \centering
    \renewcommand{\arraystretch}{1.15}
    \resizetable[0.99]{
    \begin{tabular}{llccccc}
        \toprule
        Matched View & Mismatched View  & MVTec-AD & MVTec-LOCO & VisA & GoodsAD & Average \\
        \midrule
        Full image & Full image
            & \textbf{86.25} & 67.25 & 75.76 & 69.43 & 74.67 \\
        \midrule
        \multirow{3}{*}{Matched crop}
            & No image & 85.91 & 67.72 & 74.58 & 72.15 & 75.09 \\
            & Random crop & 85.91 & 68.94 & 76.25 & 72.13 & 75.80 \\
            & Mismatched crop (ours) & 85.86 & \textbf{71.07} & \textbf{77.78} & \textbf{74.52} & \textbf{77.31} \\
        \bottomrule
    \end{tabular}
    }
\end{table}

\runinheading{Contrastive Reference Type.}
The last row of Table~\ref{tab:construction} shows how the type of contrastive reference affects calibration. Adding a random crop to the matched regional view raises the average from 75.09\% to 75.80\%. Replacing the random crop with the intended mismatched crop further increases the average to 77.31\% and improves MVTec-LOCO, VisA, and GoodsAD while leaving MVTec-AD essentially unchanged. The targeted control therefore provides more useful calibration than either no second view or a generic crop, consistent with the claim that the weight should depend on the validity of the query-reference relation.

\begin{table}[t]
    \caption{\textbf{Effect of student--teacher configurations.} We compare different student capacities and teacher constructions on the MMAD subsets. Green rows report absolute accuracy gains over the corresponding Qwen3-VL backbone in the zero-shot setting.}
    \label{tab:model_scale_ablation}
    \centering
    \renewcommand{\arraystretch}{0.97}
    \resizetable[0.82]{
        \begin{tabular}{cc*{5}{c}}
            \toprule
            Student & Teacher & MVTec-AD & MVTec-LOCO & VisA & GoodsAD & Average \\
            \midrule
            2B & 32B                                              & 77.95 & 61.92 & 68.68 & 66.42 & 68.74 \\
            \multicolumn{2}{c}{\textit{$\Delta$ vs backbone}} & \deltatext{+0.44} & \deltatext{+1.22} & \deltatext{+1.31} & \deltatext{+0.67} & \deltatext{+0.91} \\
            \midrule[0.5pt]
            4B & 32B                                              & 85.86 & 71.07 & 77.78 & 74.52 & 77.31 \\
            \multicolumn{2}{c}{\textit{$\Delta$ vs backbone}} & \deltatext{+2.69} & \deltatext{+10.80} & \deltatext{+6.65} & \deltatext{+4.41} & \deltatext{+6.14} \\
            \midrule[0.5pt]
            4B & EMA                                              & 86.10 & 66.49 & 75.23 & 72.04 & 74.96 \\
            \multicolumn{2}{c}{\textit{$\Delta$ vs backbone}} & \deltatext{+2.93} & \deltatext{+6.22} & \deltatext{+4.10} & \deltatext{+1.93} & \deltatext{+3.79} \\
            \bottomrule
        \end{tabular}
    }
\end{table}

\runinheading{Student Capacity and Teacher Construction.}
\label{sec:student_teacher_ablation}
Table~\ref{tab:model_scale_ablation} examines the roles of student capacity and teacher construction. With the same 32B teacher, the 2B student improves over its backbone by 0.91 points, whereas the 4B student improves by 6.14 points. Both students improve on every constituent dataset, but the larger gain suggests that the 4B student absorbs the privileged teacher distribution more effectively. We also compare the frozen external teacher with on-policy self-distillation (OPSD), in which the teacher is an exponential moving average (EMA) copy of the student itself. Specifically, the teacher parameters are updated as $\bar{\boldsymbol{\theta}}_t =\beta\bar{\boldsymbol{\theta}}_{t-1}+(1-\beta)\boldsymbol{\theta}_t$, where $\beta$ is the decay coefficient, set to $0.95$. The EMA teacher improves the 4B backbone by 3.79 points. Our method achieves improvements across all datasets under different student capacities and teacher constructions, showing the effectiveness of our approach.

\subsection{Qualitative Analysis}

\begin{figure}[t]
    \centering
    \includegraphics[width=\columnwidth]{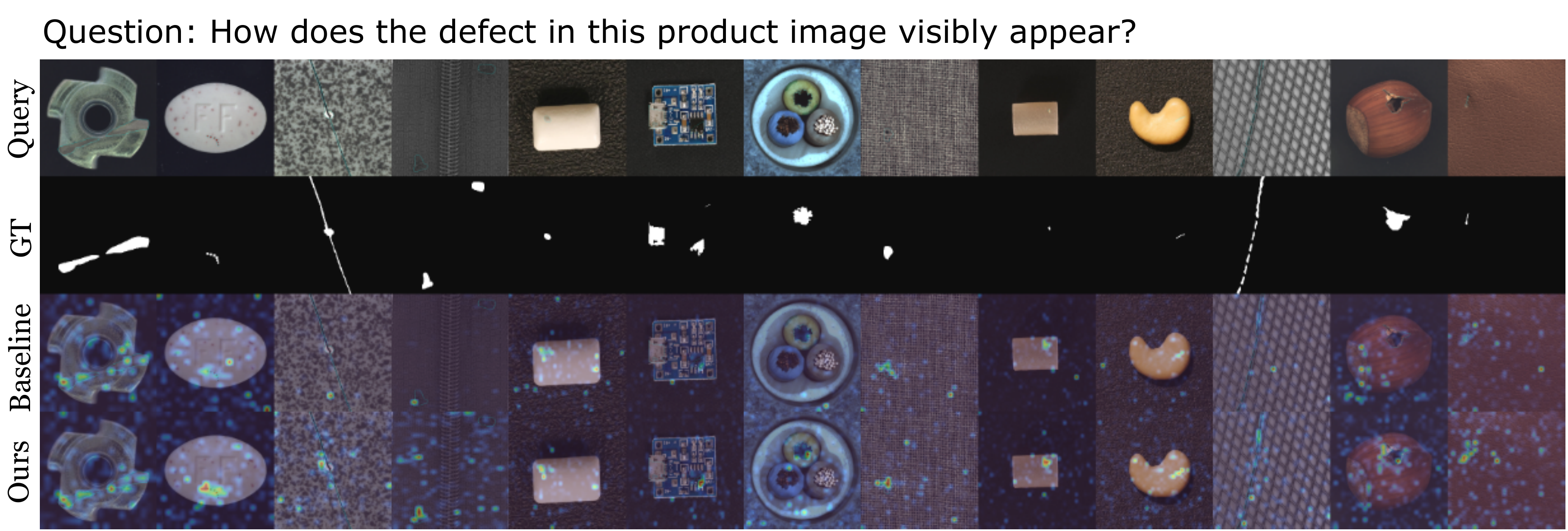}
    \caption{\textbf{Qualitative comparison of defect attribution.} Each column shows a query image, its ground-truth (GT) mask, and gradient-based attribution maps for the baseline and ADOPD.}
    \label{fig:experiment_3}
\end{figure}

Figure~\ref{fig:experiment_3} visualizes response quality by comparing gradient-based attribution maps for the baseline and ADOPD under the same prompt without references. The baseline frequently distributes activation across object interiors and surrounding textures. In contrast, ADOPD produces stronger responses near annotated defects, including thin scratches, isolated spots, and localized structural damage. This pattern is consistent with greater sensitivity to defect-relevant visual evidence and with the improvements in anomaly discrimination and defect localization reported in Table~\ref{table:0shot}. The attribution maps also show that ADOPD's responses remain broader than the ground-truth masks in several texture-heavy examples. ADOPD should therefore be viewed as improving localization-related reasoning rather than replacing a dedicated pixel-level segmentation model.

\section{Related Work}
\label{sec:related_work}

\subsection{MLLM-based Industrial Anomaly Detection}
MLLM-based IAD extends conventional anomaly scores with open-ended recognition and explanation. AnomalyGPT~\citep{anomalygpt} adapts a large vision-language model to industrial features from external vision encoders. Recent methods such as AnomalyR1~\citep{anomalyr1} and IAD-R1~\citep{iadr1} improve domain reasoning through reinforcement learning via verified rewards during post-training. AgentIAD~\citep{agentiad} utilizes external visual tools to enable agentic anomaly reasoning. Reason-IAD~\citep{reasoniad} introduces a training-free method with structured latent reasoning. These approaches strengthen industrial perception and reasoning, but they differ in training paradigm, deployment assumptions, or use of tools. ADOPD studies how reference comparison can serve as privileged information during training and be amortized into the model parameters.

\subsection{On-Policy Distillation}
On-policy distillation aligns a student with teacher feedback on student-generated states, avoiding the prefix mismatch of fixed teacher trajectories~\citep{thinkingmachine,revisiting}. Recent work views the teacher--student log-ratio as a reward and improves OPD by adjusting the reward--KL trade-off~\citep{gopd,reopold}. Complementary self-distillation methods transfer privileged information, textual feedback, or update magnitudes from a self-teacher~\citep{privileged,sdpo,rlsd}. Recent work~\citep{visionopd,vzero,vcsd} extends OPD to the multimodal reasoning field, using images as privileged information to enable fine-grained visual reasoning. OPD-IAD~\citep{opdiad} applies OPD to industrial anomaly detection, but it aims to convert generated judgments into dense anomaly maps through language-guided visual anchoring. ADOPD studies whether reference comparison can be used as privileged information for OPD in industrial reasoning.


\section{Conclusion}
\label{sec:conclusion}

In this work, we present ADOPD, a reference-privileged on-policy distillation framework that internalizes reference comparison into query-only MLLMs for industrial anomaly detection. Our central insight is that references need not remain deployment-time inputs: when used as privileged training information, they can provide both a candidate distillation direction and a reliability signal for that direction. ADOPD operationalizes this idea by evaluating student-generated rollouts with a matched reference and a mismatched control, thereby separating what the student should learn from how strongly each rollout should shape the update. The results suggest that contrastive reference supervision improves comparison-sensitive inspection behavior rather than merely adding generic visual supervision. The current dependence on paired images and defect annotations points to future work on weaker and automatically constructed privileged references.

\bibliography{adopd}

@inproceedings{mmad,
  title={MMAD: A Comprehensive Benchmark for Multimodal Large Language Models in Industrial Anomaly Detection},
  author={Jiang, Xi and Li, Jian and Deng, Hanqiu and Liu, Yong and Gao, Bin-Bin and Zhou, Yifeng and Li, Jialin and Wang, Chengjie and Zheng, Feng},
  booktitle={The Thirteenth International Conference on Learning Representations},
  year={2025}
}

@inproceedings{mvtec,
  title={MVTec AD--A comprehensive real-world dataset for unsupervised anomaly detection},
  author={Bergmann, Paul and Fauser, Michael and Sattlegger, David and Steger, Carsten},
  booktitle={Proceedings of the IEEE/CVF conference on computer vision and pattern recognition},
  pages={9592--9600},
  year={2019}
}

@article{MVTECLOCO,
  title={Beyond dents and scratches: Logical constraints in unsupervised anomaly detection and localization},
  author={Bergmann, Paul and Batzner, Kilian and Fauser, Michael and Sattlegger, David and Steger, Carsten},
  journal={International Journal of Computer Vision},
  volume={130},
  number={4},
  pages={947--969},
  year={2022},
  publisher={Springer}
}

@inproceedings{visa,
  title={Spot-the-difference self-supervised pre-training for anomaly detection and segmentation},
  author={Zou, Yang and Jeong, Jongheon and Pemula, Latha and Zhang, Dongqing and Dabeer, Onkar},
  booktitle={European conference on computer vision},
  pages={392--408},
  year={2022},
  organization={Springer}
}

@article{goodsad,
  title={PKU-GoodsAD: A supermarket goods dataset for unsupervised anomaly detection and segmentation},
  author={Zhang, Jian and Ding, Runwei and Ban, Miaoju and Dai, Linhui},
  journal={IEEE Robotics and Automation Letters},
  volume={9},
  number={3},
  pages={2008--2015},
  year={2024},
  publisher={IEEE}
}

@inproceedings{realiad,
  title={Real-IAD: A Real-World Multi-View Dataset for Benchmarking Versatile Industrial Anomaly Detection},
  author={Wang, Chengjie and Zhu, Wenbing and Gao, Bin-Bin and Gan, Zhenye and Zhang, Jiangning and Gu, Zhihao and Qian, Shuguang and Chen, Mingang and Ma, Lizhuang},
  booktitle={2024 IEEE/CVF Conference on Computer Vision and Pattern Recognition (CVPR)},
  pages={22883--22892},
  year={2024},
  organization={IEEE Computer Society}
}

@inproceedings{anomalygpt,
  title={Anomalygpt: Detecting industrial anomalies using large vision-language models},
  author={Gu, Zhaopeng and Zhu, Bingke and Zhu, Guibo and Chen, Yingying and Tang, Ming and Wang, Jinqiao},
  booktitle={Proceedings of the AAAI conference on artificial intelligence},
  volume={38},
  pages={1932--1940},
  year={2024}
}

@article{anomalyr1,
  title={Anomalyr1: A grpo-based end-to-end mllm for industrial anomaly detection},
  author={Chao, Yuhao and Liu, Jie and Tang, Jie and Wu, Gangshan},
  journal={arXiv preprint arXiv:2504.11914},
  year={2025}
}

@inproceedings{iadr1,
  title={Iad-r1: Reinforcing consistent reasoning in industrial anomaly detection},
  author={Li, Yanhui and Cao, Yunkang and Liu, Chengliang and Xiong, Yuan and Dong, Xinghui and Huang, Chao},
  booktitle={Proceedings of the AAAI Conference on Artificial Intelligence},
  volume={40},
  pages={6583--6591},
  year={2026}
}

@article{reasoniad,
  title={Towards Explainable Industrial Anomaly Detection via Knowledge-Guided Latent Reasoning},
  author={Chen, Peng and Huang, Chao and Cao, Yunkang and Liu, Chengliang and Wang, Wei and Wang, Wenqiang and Yang, Mingbo and Shen, Li and Ren, Wenqi and Cao, Xiaochun},
  journal={arXiv preprint arXiv:2602.09850},
  year={2026}
}

@inproceedings{moead,
  title={Moead: A parameter-efficient model for multi-class anomaly detection},
  author={Meng, Shiyuan and Meng, Wenchao and Zhou, Qihang and Li, Shizhong and Hou, Weiye and He, Shibo},
  booktitle={European Conference on Computer Vision},
  pages={345--361},
  year={2024},
  organization={Springer}
}

@inproceedings{anomalyclip,
  title={Anomalyclip: Object-agnostic prompt learning for zero-shot anomaly detection},
  author={Zhou, Qihang and Pang, Guansong and Tian, Yu and He, Shibo and Chen, Jiming},
  booktitle={International Conference on Learning Representations},
  volume={2024},
  pages={49705--49737},
  year={2024}
}

@article{agentiad,
  title={AgentIAD: Tool-Augmented Single-Agent for Industrial Anomaly Detection},
  author={Miao, Junwen and Du, Penghui and Liu, Yi and Wang, Yu and Wang, Yan},
  journal={arXiv preprint arXiv:2512.13671},
  year={2025}
}

@inproceedings{patchcore,
  title={Towards total recall in industrial anomaly detection},
  author={Roth, Karsten and Pemula, Latha and Zepeda, Joaquin and Sch{\"o}lkopf, Bernhard and Brox, Thomas and Gehler, Peter},
  booktitle={Proceedings of the IEEE/CVF conference on computer vision and pattern recognition},
  pages={14318--14328},
  year={2022}
}

@article{adcopilot,
  title={AD-Copilot: A Vision-Language Assistant for Industrial Anomaly Detection via Visual In-context Comparison},
  author={Jiang, Xi and Guo, Yue and Li, Jian and Liu, Yong and Gao, Bin-Bin and Deng, Hanqiu and Liu, Jun and Zhao, Heng and Wang, Chengjie and Zheng, Feng},
  journal={arXiv preprint arXiv:2603.13779},
  year={2026}
}

@article{privileged,
  title={Privileged Information Distillation for Language Models},
  author={Penaloza, Emiliano and Vattikonda, Dheeraj and Gontier, Nicolas and Lacoste, Alexandre and Charlin, Laurent and Caccia, Massimo},
  journal={arXiv preprint arXiv:2602.04942},
  year={2026}
}

@article{thinkingmachine,
  author = {Lu, Kevin and Thinking Machines Lab},
  title = {On-Policy Distillation},
  journal = {Thinking Machines Lab: Connectionism},
  year = {2025},
  doi = {10.64434/tml.20251026},
  url = {https://thinkingmachines.ai/blog/on-policy-distillation/}
}

@article{revisiting,
  title={Revisiting on-policy distillation: Empirical failure modes and simple fixes},
  author={Fu, Yuqian and Huang, Haohuan and Jiang, Kaiwen and Liu, Jiacai and Jiang, Zhuo and Zhu, Yuanheng and Zhao, Dongbin},
  journal={arXiv preprint arXiv:2603.25562},
  year={2026}
}

@misc{gopd,
  author = {Yang, Wenkai and Liu, Weijie and Xie, Ruobing and Yang, Kai and Yang, Saiyong and Lin, Yankai},
  title = {Learning beyond Teacher: Generalized On-Policy Distillation with Reward Extrapolation},
  year = {2026},
  eprint = {2602.12125},
  archivePrefix = {arXiv},
  primaryClass = {cs.LG}
}

@misc{sdpo,
  author = {H{\"u}botter, Jonas and L{\"u}beck, Frederike and Behric, Lejs and Baumann, Anton and Bagatella, Marco and Marta, Daniel and Hakimi, Ido and Shenfeld, Idan and Kleine Buening, Thomas and Guestrin, Carlos and Krause, Andreas},
  title = {Reinforcement Learning via Self-Distillation},
  year = {2026},
  eprint = {2601.20802},
  archivePrefix = {arXiv},
  primaryClass = {cs.LG}
}

@misc{vzero,
  author = {Sun, Haoxiang and Yi, Zhihang and Deng, Langxuan and Zhou, Yuhao and Jia, Peiqi and Zhao, Jian and Yuan, Li and Lv, Jiancheng and Wang, Tao},
  title = {{V-Zero}: Answer-Label-Free On-Policy Distillation with Contrastive Evidence Gating for Fine-Grained Visual Reasoning},
  year = {2026},
  eprint = {2606.25319},
  archivePrefix = {arXiv},
  primaryClass = {cs.CV}
}

@article{visionopd,
  title={Vision-opd: Learning to see fine details for multimodal llms via on-policy self-distillation},
  author={Yuan, Qianhao and Lou, Jie and Yu, Xing and Lin, Hongyu and Sun, Le and Han, Xianpei and Lu, Yaojie},
  journal={arXiv preprint arXiv:2605.18740},
  year={2026}
}

@article{rlsd,
  title={Self-distilled rlvr},
  author={Yang, Chenxu and Qin, Chuanyu and Si, Qingyi and Chen, Minghui and Gu, Naibin and Yao, Dingyu and Lin, Zheng and Wang, Weiping and Wang, Jiaqi and Duan, Nan},
  journal={arXiv preprint arXiv:2604.03128},
  year={2026}
}

@misc{reopold,
  author = {Ko, Jongwoo and Abdali, Sara and Kim, Young Jin and Chen, Tianyi and Cameron, Pashmina},
  title = {Scaling Reasoning Efficiently via Relaxed On-Policy Distillation},
  year = {2026},
  eprint = {2603.11137},
  archivePrefix = {arXiv},
  primaryClass = {cs.LG}
}

@article{vcsd,
  title={Visual Contrastive Self-Distillation},
  author={Liang, Yijun and Tian, Yunjie and Li, Yijiang and Jia, Yuqi and Huang, Furong and Zhou, Tianyi and Fu, Di},
  journal={arXiv preprint arXiv:2607.21556},
  year={2026}
}

@article{opdiad,
  title={OPD-IAD: From Language Judgment to Industrial Anomaly Detection via On-Policy Self-Distillation},
  author={Chen, Shuimu and Jin, Jing and Su, Nan and Xu, Hongbo and Cheng, Zebang and Yang, Wenming and Ma, Fei and Wang, Guijin},
  journal={arXiv preprint arXiv:2607.18850},
  year={2026}
}

@article{qwen3vl,
  title={Qwen3-vl technical report},
  author={Bai, Shuai and Cai, Yuxuan and Chen, Ruizhe and Chen, Keqin and Chen, Xionghui and Cheng, Zesen and Deng, Lianghao and Ding, Wei and Gao, Chang and Ge, Chunjiang and others},
  journal={arXiv preprint arXiv:2511.21631},
  year={2025}
}

@article{qwen2.5vl,
  title={Qwen2.5-VL Technical Report},
  author={Bai, Shuai and Chen, Keqin and Liu, Xuejing and Wang, Jialin and Ge, Wenbin and Song, Sibo and Dang, Kai and Wang, Peng and Wang, Shijie and Tang, Jun and others},
  journal={arXiv preprint arXiv:2502.13923},
  year={2025}
}

@inproceedings{llava,
  title={Improved baselines with visual instruction tuning},
  author={Liu, Haotian and Li, Chunyuan and Li, Yuheng and Lee, Yong Jae},
  booktitle={Proceedings of the IEEE/CVF conference on computer vision and pattern recognition},
  pages={26296--26306},
  year={2024}
}

@misc{llavanext,
    title={LLaVA-NeXT: Improved reasoning, OCR, and world knowledge},
    url={https://llava-vl.github.io/blog/2024-01-30-llava-next/},
    author={Liu, Haotian and Li, Chunyuan and Li, Yuheng and Li, Bo and Zhang, Yuanhan and Shen, Sheng and Lee, Yong Jae},
    month={January},
    year={2024}
}

@article{glm,
  title={Glm-4.5 v and glm-4.1 v-thinking: Towards versatile multimodal reasoning with scalable reinforcement learning},
  author={Hong, Wenyi and Yu, Wenmeng and Gu, Xiaotao and Wang, Guo and Gan, Guobing and Tang, Haomiao and Cheng, Jiale and Qi, Ji and Ji, Junhui and Pan, Lihang and others},
  journal={arXiv preprint arXiv:2507.01006},
  year={2025}
}

@article{internvl3,
  title={Internvl3: Exploring advanced training and test-time recipes for open-source multimodal models},
  author={Zhu, Jinguo and Wang, Weiyun and Chen, Zhe and Liu, Zhaoyang and Ye, Shenglong and Gu, Lixin and Tian, Hao and Duan, Yuchen and Su, Weijie and Shao, Jie and others},
  journal={arXiv preprint arXiv:2504.10479},
  year={2025}
}

@article{hybridflow,
  title   = {HybridFlow: A Flexible and Efficient RLHF Framework},
  author  = {Guangming Sheng and Chi Zhang and Zilingfeng Ye and Xibin Wu and Wang Zhang and Ru Zhang and Yanghua Peng and Haibin Lin and Chuan Wu},
  year    = {2024},
  journal = {arXiv preprint arXiv: 2409.19256}
}
\bibliographystyle{iclr2027_conference}

\clearpage


\appendix

\section{On-Policy Distillation Derivation}
\label{app:opd_derivation}
Here, we derive the expected gradients of the OPD objective introduced in Section~\ref{sec:preliminary}. The derivation follows the notation of~\citep{gopd} and is adapted to the multimodal setting. Let $h_t=(x,y_{<t})$ denote a prompt together with a student-generated prefix, and let $\pi_S(\cdot\mid h_t)$ and $\pi_T(\cdot\mid h_t)$ be the student and teacher next-token distributions, respectively. Specifically, $\theta$ denotes the parameters of the student model. In sampled-token OPD, the prefix is treated as a stop-gradient training state.

\paragraph{Fixed-prefix reverse KL.}
At a fixed state $h_t$, OPD minimizes the local reverse KL divergence
\begin{equation}
    \mathcal{J}(h_t)=D_{\mathrm{KL}}\left(\pi_S(\cdot\mid h_t)\parallel\pi_T(\cdot\mid h_t)\right) =\sum_v \pi_S(v\mid h_t)\log\frac{\pi_S(v\mid h_t)}{\pi_T(v\mid h_t)}.
    \label{eq:app_local_reverse_kl}
\end{equation}
Here, $v$ ranges over the vocabulary. For compactness, define the local log-ratio $\delta_t(v)=\log\frac{\pi_S(v\mid h_t)}{\pi_T(v\mid h_t)}$. Because the teacher is frozen, $\nabla_{\theta}\delta_t(v)=\nabla_{\theta}\log\pi_S(v\mid h_t)$. Applying the product rule to each vocabulary term gives
\begin{equation}
    \begin{aligned}
        \nabla_{\theta}\!\left[\pi_S(v\mid h_t)\delta_t(v)\right]
        &=\nabla_{\theta}\pi_S(v\mid h_t)\delta_t(v)+\pi_S(v\mid h_t)\nabla_{\theta}\delta_t(v) \\
        &=\pi_S(v\mid h_t)\nabla_{\theta}\log\pi_S(v\mid h_t)\delta_t(v)+\pi_S(v\mid h_t)\nabla_{\theta}\log\pi_S(v\mid h_t) \\
        &=\pi_S(v\mid h_t)\left(\delta_t(v)+1\right)\nabla_{\theta}\log\pi_S(v\mid h_t).
    \end{aligned}
    \label{eq:app_product_rule_local_kl}
\end{equation}
Because Eq.~\eqref{eq:app_product_rule_local_kl} gives the gradient of a single vocabulary term, analytically summing it over all possible next tokens recovers the exact fixed-prefix reverse-KL gradient:
\begin{equation}
    \nabla_{\theta} \mathcal{J}(h_t) =\mathbb{E}_{v\sim\pi_S(\cdot\mid h_t)} \left[ \left(\delta_t(v)+1\right)\nabla_{\theta}\log\pi_S(v\mid h_t)\right].
    \label{eq:app_local_gradient_with_one}
\end{equation}
The additive score term is zero:
\begin{equation}
    \mathbb{E}_{v\sim\pi_S}\left[\nabla_{\theta}\log\pi_S(v\mid h_t)\right]=\sum_v \nabla_{\theta}\pi_S(v\mid h_t)=\nabla_{\theta}\sum_v\pi_S(v\mid h_t)=\nabla_{\theta} 1=0.
    \label{eq:app_score_identity}
\end{equation}
Removing this zero-mean term gives the local reverse-KL gradient used by OPD:
\begin{equation}
    \nabla_{\theta} \mathcal{J}(h_t)=\mathbb{E}_{v\sim\pi_S(\cdot\mid h_t)}\left[\delta_t(v)\nabla_{\theta}\log\pi_S(v\mid h_t)\right].
    \label{eq:app_local_reverse_kl_gradient}
\end{equation}

\paragraph{Sampled-token estimator and OPD advantage.}
For a token $y_t\sim\pi_S(\cdot\mid h_t)$, the following quantity is an unbiased Monte Carlo estimator of the fixed-prefix gradient in Eq.~\eqref{eq:app_local_reverse_kl_gradient}:
\begin{equation}
    \widehat{\nabla_{\theta} \mathcal{J}}=\delta_t(y_t)\nabla_{\theta}\log\pi_S(y_t\mid h_t).
    \label{eq:app_sampled_opd_gradient}
\end{equation}
Indeed, taking the expectation over the sampled token recovers Eq.~\eqref{eq:app_local_reverse_kl_gradient} exactly:
\begin{equation}
    \mathbb{E}_{y_t\sim\pi_S}\left[\widehat{\nabla_{\theta} \mathcal{J}}\right]=\sum_v \pi_S(v\mid h_t)\delta_t(v)\nabla_{\theta}\log\pi_S(v\mid h_t) =\nabla_{\theta}\mathcal{J}(h_t).
    \label{eq:app_mc_unbiasedness}
\end{equation}
Equivalently, defining the detached token-level advantage as
\begin{equation}
    A_t=\operatorname{sg}\!\left[\log\pi_T(y_t\mid h_t)-\log\pi_S(y_t\mid h_t)\right]
    \label{eq:app_opd_advantage}
\end{equation}
gives the stop-gradient surrogate
\begin{equation}
    \widehat{\mathcal{L}}_{\mathrm{OPD}}=-\frac{1}{T}\sum_{t=1}^{T}A_t\log\pi_S(y_t\mid h_t).
    \label{eq:app_opd_surrogate}
\end{equation}
Since $A_t$ is detached, the gradient of one surrogate term is
\begin{equation}
    \nabla_{\theta}\left[-A_t\log\pi_S(y_t\mid h_t)\right]=\delta_t(y_t)\nabla_{\theta}\log\pi_S(y_t\mid h_t),
    \label{eq:app_surrogate_gradient}
\end{equation}
which is exactly Eq.~\eqref{eq:app_sampled_opd_gradient}. Thus, a sampled token is reinforced when the teacher assigns it higher probability than the student and suppressed when the student overweights it.

\section{Dataset Construction}
\label{app:realiad_paired_dataset}

\paragraph{Dataset source.}
Real-IAD~\citep{realiad} is a large-scale real-world dataset for industrial anomaly detection. It includes 30 object classes, each with five shooting angles, and contains 150K high-resolution images. We sample paired subsets of Real-IAD to construct a training set for ADOPD. Each pair consists of one anomalous image and its corresponding normal counterfactual image. The paired dataset is used to generate matched and mismatched privileged references for the teacher during training. Our final dataset contains 6,000 training records.

\paragraph{Dataset processing.}
An anomalous sample is retained only when its segmentation mask contains a valid connected component of at least 16 pixels. Masks are binarized using a threshold of 128. Nearby components are merged, and at most four defect regions are retained. To improve instance diversity, no physical instance contributes more than two camera views. The student receives the original $1024\times1024$ inspection image without a defect bounding box. In contrast, teacher supervision uses a localized crop in which the target region is indicated by a red bounding box. Crop windows include surrounding context, have a minimum side length of 128 pixels, and are resized such that the short side is 336 pixels. Multiple separated defect regions are organized as a montage.

\paragraph{Prompt construction and storage format.}
Each record contains one student image, one matched teacher image, and one mismatched teacher image. The serialized
schema consists of \texttt{prompt}, \texttt{image}, \texttt{teacher\_raw\_prompt}, \texttt{teacher\_image}, and \texttt{teacher\_neg\_image}. The ground-truth answer is retained in the audit metadata but is not exposed in the training record. An example of the serialized prompt is shown in Table~\ref{tab:dataset_construction_overview}.

\begin{table}[tbhp]
    \caption{Example prompt used for ADOPD.}
    \label{tab:dataset_construction_overview}
    \centering
    \resizetable{%
        \begin{tabular}{@{}p{\linewidth}@{}}
            \toprule
            Student Input\; \texttt{prompt}: \\
            \texttt{\textless image\textgreater} How should the visible defect in this product image be classified?\\
            A. deformation B. contamination C. physical damage D. pit \\
            \\
            Teacher Input\; \texttt{teacher\_raw\_prompt}: \\
            \texttt{\textless teacher\_image\textgreater} \texttt{\textless teacher\_neg\_image\textgreater} You can refer to the regional reference image to answer the question. \\
            \bottomrule
        \end{tabular}
    }
\end{table}

\end{document}